\UseRawInputEncoding
\pdfoutput=1
\documentclass[letterpaper, 10 pt, conference]{ieeeconf}  

\IEEEoverridecommandlockouts                              

\usepackage{array}
\usepackage{amsmath} 
\usepackage{amssymb}  
\usepackage{graphicx}
\usepackage{url}
\usepackage{multirow}
\usepackage[ruled, vlined, linesnumbered]{algorithm2e}
\usepackage[colorlinks,linkcolor=blue,anchorcolor=blue]{hyperref}
\usepackage[utf8]{inputenc}

\title{\LARGE \bf
Unsupervised Domain Adaptation for Point Cloud Semantic Segmentation via Graph Matching
}

\author{Yikai Bian, Le Hui, Jianjun Qian$^{*}$ and Jin Xie$^{*}$
	\thanks{PCA Lab, Key Lab of Intelligent Perception and Systems for High-Dimensional Information of Ministry of Education, and Jiangsu Key Lab of Image and Video Understanding for Social Security, School of Computer Science and Engineering, Nanjing University of Science and Technology, Nanjing, China. Email:\{yikai.bian, le.hui, csjqian, csjxie\}@njust.edu.cn.}
	\thanks{$^{*}$The author responsible for the correspondence of this paper.}%
}

\begin{document}

\maketitle
\thispagestyle{empty}
\pagestyle{empty}

\begin{abstract}

Unsupervised domain adaptation for point cloud semantic segmentation has attracted great attention due to its effectiveness in learning with unlabeled data. Most of existing methods use global-level feature alignment to transfer the knowledge from the source domain to the target domain, which may cause the semantic ambiguity of the feature space. In this paper, we propose a graph-based framework to explore the local-level feature alignment between the two domains, which can reserve semantic discrimination during adaptation. Specifically, in order to extract local-level features, we first dynamically construct local feature graphs on both domains and build a memory bank with the graphs from the source domain. In particular, we use optimal transport to generate the graph matching pairs. Then, based on the assignment matrix, we can align the feature distributions between the two domains with the graph-based local feature loss. Furthermore, we consider the correlation between the features of different categories and formulate a category-guided contrastive loss to guide the segmentation model to learn discriminative features on the target domain. Extensive experiments on different synthetic-to-real and real-to-real domain adaptation scenarios demonstrate that our method can achieve state-of-the-art performance. Our code is available at \url{https://github.com/BianYikai/PointUDA}.

\end{abstract}

\section{INTRODUCTION}

Deep learning methods \cite{qi2017pointnet,wang2019graph} for point cloud semantic segmentation have shown dramatic success in recent years. However, most of these methods focus on fully supervised learning for point cloud segmentation with a large number of manually annotated labels. Although there are several public datasets providing large amounts of annotation data, it is difficult to directly apply the model trained on a labeled source domain to another unlabeled target domain. The reason lies in that the data collected by different 3D sensors have a huge discrepancy in appearance and sparsity, which results in the domain shift problem. Therefore, how to generalize a well-trained model to another unlabeled domain is a challenging but valuable problem in point cloud semantic segmentation.

Unsupervised domain adaptation can alleviate the domain shift problem by transferring the knowledge from the labeled source domain to the unlabeled target domain. Recent advances on unsupervised point cloud domain adaptation tasks mainly focus on reducing the domain gap between the inputs. For example, Yi \textit{et al.} \cite{yi2021complete} build a point cloud completion network with sequences of point clouds to bridge  the domain gap between LiDAR sensors with different beams. ePointDA \cite{zhao2021epointda} and SqueezeSegV2 \cite{wu2019squeezesegv2} use auxiliary rendering networks to render dropout noises or intensity on the synthetic dataset, which translate the point clouds from the source domain similar to the target domain. Furthermore, these methods use a series of feature alignment methods to increase the consistency of feature distributions, such as higher-order moment matching \cite{chen2020homm}, and geodesic correlation alignment \cite{morerio2018minimal}. However, these methods mainly consider the overall distributions of two domains to form the global-level feature alignment, which ignores the local geometric differences between the domains.

In this paper, we propose a domain adaptation framework for unsupervised point cloud segmentation with the local-level feature alignment. Compared with the global-level feature alignment, our framework can focus on the correlation between the similar local structures of point from the two domains, so that the reliable feature alignment can be performed to guide the discriminative semantic feature learning of the target domain. Specifically, through the farthest point sampling, we select a set of centroid points and construct the dynamic local feature graph for each centroid point to capture its local geometry information. Then, in order to enrich the graph of the source domain, we construct a feature graph memory bank to store the generated source-domain feature graphs during the training phase. After that, inspired by the point cloud matching \cite{yew2020-RPMNet}, we adopt the optimal-transport cost to measure the graph similarities between the memory bank and target domain, so that a reliable assignment matrix can be obtained to guide the knowledge transferring from the source domain to the target domain. Particularly, in order to further extract the discriminative target-domain feature, we consider the category-wise correlation between the source domain and the target domain, and exploit the contrastive learning to increase category-level discrimination of target graphs. Such category-guided contrastive loss can effectively help cluster and distinguish the feature-graph distributions of different categories. Extensive experiments demonstrate the effectiveness of our framework, where we not only focus on the synthetic-to-real domain adaptation scenarios (vKITTI to SemanticPOSS), but also pay attention to the indoor (S3DIS to ScanNet) and the outdoor (SemanticKITTI to nuScenes) real-to-real domain adaptation scenarios. 

Our contributions can be summarized as follows:
\begin{itemize}
	\item We propose a novel graph-based framework for local-level feature alignment for unsupervised domain adaptive point cloud semantic segmentation.
	\item We construct feature graphs to capture the local geometry information of point clouds and use a local feature loss based on an assignment matrix for the alignment of feature graphs.
	\item We develop a category-guided contrastive loss to guide the segmentation model to learn the discriminative features on the target domain.
\end{itemize}

\section{RELATED WORK}

\textbf{Point Cloud Semantic Segmentation}. Recent progress on point cloud semantic segmentation is mainly divided into several categories according to different representations of data. Volumetric-based methods require a preprocessing stage to voxelize the original point cloud. SparseConvNet \cite{graham20183d} proposes a submanifold sparse convolution network to deal with spatially-sparse voxel data. MinkowskiNet \cite{choy20194d} creates Minkowski space on sparse representation data and proposes a powerful 4-dimensional convolutional neural network to deal with 3D videos. Projection-based methods need to project the point cloud into an image before feeding data into the network. SqueezeSegV2 \cite{wu2019squeezesegv2} uses a context aggregation module to improve the robustness to dropout noise on projected 2D LiDAR image. SqueezeSegV3 \cite{xu2020squeezesegv3} proposes an efficient spatially-adaptive convolution to deal with the discrepancy of data distribution of different LiDAR image locations. Point-based methods directly use unordered point clouds for semantic segmentation. However, due to the heavy computation, most of the methods first split the point cloud into blocks before training and inferring. PointNet \cite{qi2017pointnet} uses multi-layer perceptron and a mini-network (T-Net) to extract features from unordered point clouds. In order to strengthen the local information for point-level segmentation, GACNet \cite{wang2019graph} and PointWeb \cite{zhao2019pointweb} use different attention modules to dynamically assign weight to local features. In this work, we leverage PointWeb as our segmentation network because of its efficiency in processing unordered point clouds. 

\noindent\textbf{Unsupervised Domain Adaptation}. Unsupervised domain adaptation (UDA) aims to train the model in the labeled source domain and generalize the knowledge to the target domain through the unsupervised methods. Recent advances on domain adaptation for 3D point cloud mainly study aligning the distributions by input-level and feature-level alignment. Saleh \textit{et al.} \cite{saleh2019domain} use CycleGAN \cite{zhu2017unpaired} to translate the synthetic bird's eye view point cloud image to the real point cloud image for domain adaptive vehicle detection. ePointDA \cite{zhao2021epointda} use a dropout noise rendering network to achieve uniformity of data distribution between domains and adopt a higher-order moment matching loss for feature-level alignment. Yi \textit{et al.} \cite{yi2021complete} use a completion network to complete the point cloud with sequences data, so that they can recover the 3D surfaces from different LiDAR data and transfer knowledge between different LiDAR sensors. However, the input-level adaptation methods lead to extra challenges and training costs due to the variable geometric structures in different domains. Besides, recent works on 2D UDA \cite{chen2019domain,vu2019advent,tzeng2017adversarial} are also quite applicable to 3D UDA, where they use the different losses to decrease the domain shift problem, $e.g.$ maximum squares loss, entropy loss, and adversarial loss. Furthermore, self-training is also an effective technique for UDA. \cite{achituve2021self} proposes a self-supervised task for target domain to learn its useful representations. ST3D \cite{yang2021st3d} proposes a quality-aware triplet memory bank to generate high-quality 3D detection pseudo labels for self-training. xMUDA \cite{jaritz2020xmuda} and DsCML \cite{peng2021sparse} propose cross-modal constraint to retain the advantages of 2D images and 3D point clouds for domain adaptation. In this paper, considering the input-level methods cannot handle complex domain adaptation scenarios, we develop a general uni-modal 3D UDA framework with feature-level alignment.

\begin{figure*}[thpb]
	\centering
	\includegraphics[width=1.0\linewidth]{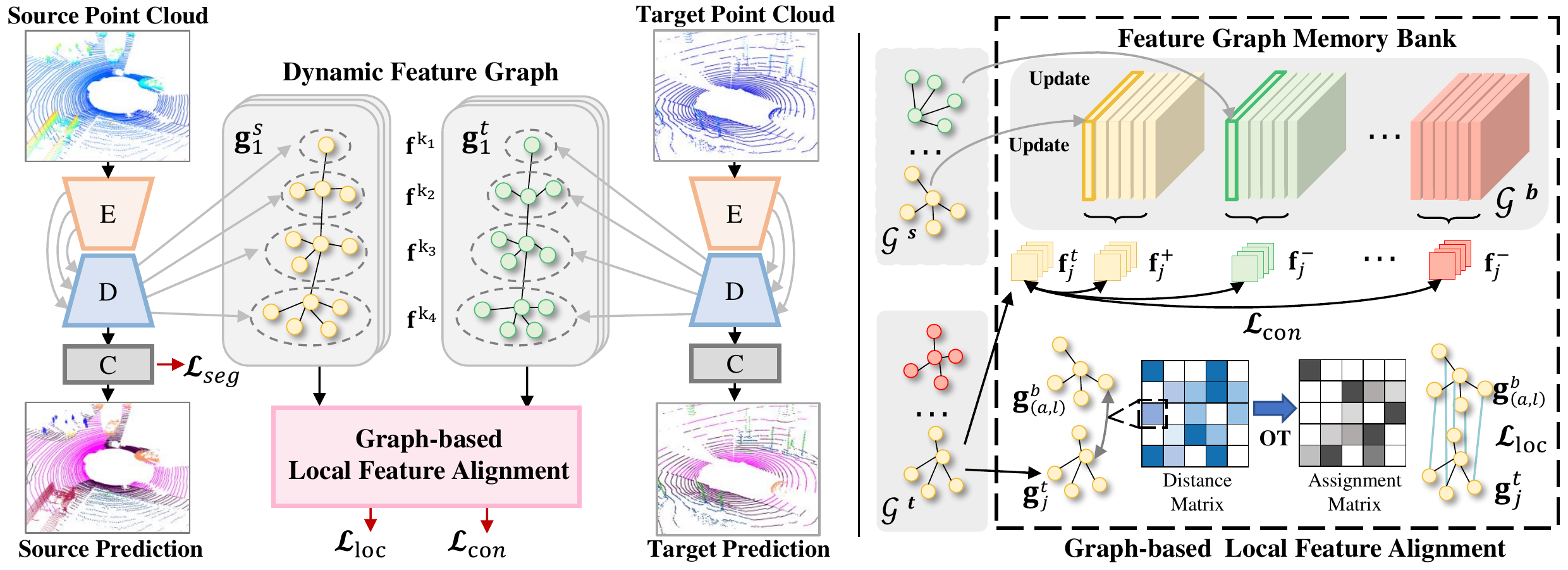}
	\vspace{-20pt}
	\caption{Illustration of our framework. Point clouds in the source and target domains are fed into the model to construct dynamic feature graphs. The source graphs are used to build a memory bank and the target graphs are aligned by our graph-based local feature alignment method.}
	\label{fig:framwork}
	\vspace{-13pt}
\end{figure*}

\section{OUR METHOD}
\subsection{Overview} 
In unsupervised domain adaptive point cloud semantic segmentation, we are able to access the source domain $\mathcal{X}_s=\{\mathbf{x}^s_i\}^{M^s}_{i=1}$ of ${M^s}$ point clouds with its segmentation labels $\mathcal{Y}_s=\{\mathbf{y}^s_i\}^{M^s}_{i=1}$ and the unlabeled target domain $\mathcal{X}_t=\{\mathbf{x}^t_j\}^{M^t}_{j=1}$ of ${M^t}$ point clouds, where $\mathbf{x}^s_i\in\mathbb{R}^{{N^s_i}\times3}$ is the set of ${N^s_i}$ points and $\mathbf{x}^t_j\in\mathbb{R}^{{N^t_j}\times3}$ is the set of ${N^t_j}$ points. Given data $\mathcal{X}_s$, $\mathcal{Y}_s$ and $\mathcal{X}_t$, our goal is to train a model which can precisely categorize the point of target data into one of the common semantic categories in the source data, and to alleviate the performance drop problem caused by the domain gap at the same time. As illustrated in Fig.\ref{fig:framwork}, we first construct the local feature distributions of the source and target domains with the proposed dynamic feature graphs. Then, by building a source-domain feature graph memory bank, we employ graph matching to obtain graph pairs between the graphs in the memory bank and the target graphs. Finally, with the obtained matching association, we utilize the designed losses for local-level feature alignment.

\subsection{Dynamic Feature Graph} 
\label{sec:frame}

Different from the global-level feature alignment methods \cite{chen2020homm,morerio2018minimal,chen2019domain,vu2019advent,tzeng2017adversarial} roughly aligning two domains, we consider the differences of local neighborhood context in the target domain for the fine alignment. The main idea of our method is to use the learned dynamic local feature graphs to capture the multi-level features in different neighborhoods of point clouds. Then, based on the local-graph similarity, the correlation between local neighborhoods from two domains can be viewed as the knowledge transferring from the labeled source domain to the unlabeled target domain.

We leverage PointWeb \cite{zhao2019pointweb} as our semantic segmentation backbone, which contains a classifier and a feature extractor with an encoder and a decoder. We use the feature extractor to extract the multi-level features and then build dynamic feature graphs on the sampled centroid points by the feature similarity. Specifically, given a point cloud sample $\mathbf{x}\in\mathbb{R}^{{N}\times3}$ with $N$ points, we first extract its local-context feature using the feature extractor. Then, we select ${N/64}$ centroid points using the farthest point sampling for three iterations, where the centroid points are then as the kernel points of graphs for feature aggregation at each level. In detail, for each kernel point at different levels, we gather its k-nearest neighbors in the feature space ($k$ is different at each level). Thereby, a feature graph can be constructed by setting the k-NN features as its vertices $\{\mathbf{v}_{i}^{k_1}, \mathbf{v}_{i}^{k_2}, \mathbf{v}_{i}^{k_3}, \mathbf{v}_{i}^{k_4}\}$ and the k-NN feature distances as the edges $\{\mathbf{e}_{i}^{k_1}, \mathbf{e}_{i}^{k_2}, \mathbf{e}_{i}^{k_3}, \mathbf{e}_{i}^{k_4}\}$, where $i$ is the index of centroid points and $\{k_j\}_{j=1}^4$ is the different values in k-NN. We obtain ${N/64}$ dynamically updated local feature graphs to represent the local neighborhood context of the given point cloud, which can be formulated as,
\begin{equation}
\label{loss:1}
\setlength{\abovedisplayskip}{7pt}
\setlength{\belowdisplayskip}{5pt}
\begin{aligned}
\mathcal{G}={\{\mathbf{g}_i\}}^{N_c}_{i=1}=\{\{\mathbf{f}_{i}^{k_j}\}^{4}_{j=1}\}^{N_c}_{i=1}=\{\{\mathbf{v}_{i}^{k_j},\mathbf{e}_{i}^{k_j}\}^{4}_{j=1}\}^{N_c}_{i=1},
\end{aligned}
\end{equation}
where $N_c$ denotes the number of ${N/64}$ centroid points, and $\mathbf{f}_{i}^{k_j}$ represents the feature embedding containing the vertice and the edge information. As a result, given a source sample and a target sample, the generated graphs are represented as $\mathcal{G}^s={\{\mathbf{g}^s_i\}}^{N_c}_{i=1}$ and $\mathcal{G}^t={\{\mathbf{g}^t_i\}}^{N_c}_{i=1}$.

\subsection{Graph-based Local Feature Alignment} 
\label{sec:align}

In this section, based on the constructed dynamic graphs above, we aim to find the intrinsic correlation between the source domain and the target domain. We use the graphs from the source domain to guide the model to extract semantic discriminative features on the target domain. However, at the training stage, the sample category in a batch is limited and their graph patterns tend to present significant structure differences, which may potentially introduce the alignment bias. To address the issue above, we build up a feature graph memory bank $\mathcal{G}^b$ and store a graph $\mathbf{g}^s_i$ into the bank according to the corresponding category of the centroid point. Therefore, benefiting from such memory bank mechanism, we can sufficiently mine the rich source information from it for reliable target-domain feature learning. The memory bank provides the same capacity $B$ for each category of graphs. Once the number of graphs exceeds the capacity $B$ of the corresponding category, we will update the memory bank by replacing the oldest graphs with the new ones.

Given the graphs $\mathcal{G}^t={\{\mathbf{g}^t_j\}}^{N_c}_{j=1}$ from target domain, we consider finding the most similar graph from $\mathcal{G}^b$ to each graph in $\mathcal{G}^t$ for feature alignment. In particular, we use optimal transport for graph matching. Specifically, the total transport cost of optimal transport is used to measure the similarity between two graphs, and the assignment matrix $\mathbf{A}\in\mathbb{R}^{K\times K}$ is used to find the point-level correspondences for $K$ nodes in graphs. In the graph matching formulation, we first compute the distance matrix $\mathbf{D}\in\mathbb{R}^{K\times K}$, where the element $\mathbf{D}_{(m,n)}$ indicates the distance between the point $m$ in one graph and the point $n$ in the other graph. Here, we use the squared Euclidean distance in the feature space to measure the pairwise distance between points in graphs, where the points are composed of the corresponding features with edge and vertice information. Once we obtain the distance matrix $\mathbf{D}$, we apply the Sinkhorn algorithm \cite{cuturi2013sinkhorn} to obtain the final assignment matrix $\mathbf{A}$ and the total transport cost through solving the optimal transport problem. In this way, we can measure the relevance between each target graph with all the graphs in the memory bank. As a result, we can find the most similar graph $\mathbf{g}^b_{(a,l)}\in\mathcal{G}^b$  for the target graph $\mathbf{g}^t_j$ according to the sorting result of the transport cost for knowledge transferring, where $a$ indicates the category and $l$ indicates the index in the memory bank.

Based on the generated graph pairs, we formulate a local feature loss based on the assignment matrix for the local-level feature alignment. Given a target graph  $\mathbf{g}^t_j$, we first select the most similar graph $\mathbf{g}^b_{(a,l)}$ from the memory bank. At the same time, we are able to access the corresponding assignment matrix $\mathbf{A}_j\in \mathbb{R}^{K\times K}$, which can decode the point-level corresponding feature assignment between two graphs. In detail, for each point in target graph  $\mathbf{g}^t_j$, we can obtain the corresponding transport weights for every point in graph $\mathbf{g}^b_{(a,l)}$ from the assignment matrix $\mathbf{A}_j$. Then, we perform a weighted sum of $\mathbf{g}^b_{(a,l)}\in\mathbb{R}^{K\times D}$ to guide the learning of $\mathbf{g}^t_j\in\mathbb{R}^{K\times D}$, where $D$ is the number of feature channels. The key point is that the local neighborhood areas with similar semantic contexts need to have similar feature distributions. In this way, we can effectively align the indiscriminate feature distributions of the unlabeled target domain to the source domain. Therefore, we propose the following assignment matrix based local feature loss for feature graph learning in the target domain:
\begin{equation}
\setlength{\abovedisplayskip}{4pt}
\setlength{\belowdisplayskip}{4pt}
\begin{aligned}
\mathcal{L}_{loc}=\frac{1}{N_c}\sum_{j=1}^{N_c}{\left\|\mathbf{g}^t_j-\mathbf{A}_j\mathbf{g}^b_{(a,l)}\right\|}_1.
\end{aligned}
\end{equation}

Owing to a variety of feature graphs from different categories in our memory bank, we can further exploit the contrastive learning for more discriminative target-domain feature learning. Here, we select the category $a$ of the matched graph $\mathbf{g}^b_{(a,l)}$ as the positive category and the other categories as the negative categories. In order to obtain the representative features of each category, all graphs in the memory bank are used for calculating the feature representations. Although we have achieved the assignment matrix for each positive or negative pair, it is meaningless for point-level adaptation on unmatched pairs. Therefore, we use the mean of all features in the graphs with the same category to represent the feature representation of the corresponding category. It is worth noting that our graph is composed of multi-level features, so we calculate the mean features of different levels separately and then concatenate them as the final feature representation. The positive and negative features for graph $\mathbf{g}^t_j$ can be formulated as,
\begin{equation}
\setlength{\abovedisplayskip}{4pt}
\setlength{\belowdisplayskip}{1pt}
\begin{aligned}
\mathbf{f}^+_j=\frac{1}{B}\sum_{b=1}^{B}\sum_{c=1}^{C}\mathbb{I}_{[c=a]}\Theta\left({\mathbf{f}_{(b,c)}^{k_1}, \mathbf{f}_{(b,c)}^{k_2}, \mathbf{f}_{(b,c)}^{k_3}, \mathbf{f}_{(b,c)}^{k_4}}\right),
\end{aligned}
\end{equation}
\begin{equation}
\setlength{\abovedisplayskip}{1pt}
\setlength{\belowdisplayskip}{4pt}
\begin{aligned}
\mathbf{f}^-_j=\frac{1}{BC}\sum_{b=1}^{B}\sum_{c=1}^{C}\mathbb{I}_{[c\neq a]}\Theta\left({\mathbf{f}_{(b,c)}^{k_1}, \mathbf{f}_{(b,c)}^{k_2}, \mathbf{f}_{(b,c)}^{k_3}, \mathbf{f}_{(b,c)}^{k_4}}\right),
\end{aligned}
\end{equation}
where $C$ is the number of categories and $B$ is the capacity size of each category. The indicator function $\mathbb{I}$ returns 1 if the condition is satisfied or returns 0 if unsatisfied. We use $\Theta$ to represent the mean and concatenation operators.

Then, with the generated positive and negative features, we formulate the following contrastive loss for increasing the intra-category compactness and inter-category separability between the target graph $\mathbf{g}^t_j$ and the graphs in $\mathcal{G}^b$.
\begin{equation}
\setlength{\abovedisplayskip}{1pt}
\setlength{\belowdisplayskip}{1pt}
\begin{aligned}
\mathcal{L}_{con}=\frac{1}{N_c}\sum_{j=1}^{N_c}\left[{\left\|\mathbf{f}^t_j-\mathbf{f}^+_j\right\|}_1 - {\left\|\mathbf{f}^t_j -\mathbf{f}^-_j\right\|}_1 + \alpha \right]_+,
\end{aligned}
\end{equation}
where $\alpha$ is the margin of the contrastive loss and $\mathbf{f}^t_j$ is the mean feature of graph $\mathbf{g}^t_j$. Therefore, with the proposed local feature loss and the contrastive loss, we consider the feature alignment from two complementary perspectives, which can significantly reduce the domain discrepancy in feature space.


\subsection{Domain Adaptation Scheme}
\label{sec:scheme}
For the unsupervised domain adaptation, the core challenge is how to learn the discriminative target features without labels. First of all, for the source domain, we use the standard cross-entropy loss for supervised training:
\begin{equation}
\setlength{\abovedisplayskip}{1pt}
\setlength{\belowdisplayskip}{1pt}
\begin{aligned}
\mathcal{L}_{seg}=-\frac{1}{N}\sum_{n=1}^{N}\sum_{c=1}^{C}\mathbf{y}^s_{(n,c)}\log\hat{\mathbf{y}}^s_{(n,c)},
\end{aligned}
\end{equation}
where $\mathbf{y}^s\in\mathbb{R}^{N\times C}$ is the semantic labels for $N$ points with $C$ semantic categories and $\hat{\mathbf{y}}^s$ is the outputs from the model.

In addition, in our framework, in order to identify the relationship of local features between the source domain and the target domain, we construct dynamic feature graphs and the generated graph pairs based on the graph matching to find correspondences between the two domains. With our developed local feature loss based on the assignment matrix and the category-guided contrastive loss, we can effectively align the local features between the two domains. The overall loss can be formulated as:
\begin{equation}
\label{loss:complete}
\setlength{\abovedisplayskip}{4pt}
\setlength{\belowdisplayskip}{4pt}
\begin{aligned}
\mathcal{L}_{all}=\mathcal{L}_{seg}+\lambda_1\mathcal{L}_{loc}+\lambda_2\mathcal{L}_{con}
\end{aligned},
\end{equation}
where $\lambda_1$ and $\lambda_2$ are hyperparameters balancing the proposed losses with semantic segmentation loss.

Furthermore, our framework can be extended into a two-stage method in a  self-training manner, where we follow Jaritz \textit{et al.} \cite{jaritz2020xmuda} to use a pseudo-label training strategy. We first use our framework to train a model with the loss in Eq. \ref{loss:complete}, where the source data $\mathcal{X}_s$, $\mathcal{Y}_s$ and the target data $\mathcal{X}_t$ are available. Then we fix the parameters of the model and generate pseudo labels $\mathcal{\hat{Y}}_s$ for target data. After that, the supervised semantic segmentation loss with pseudo labels is used on the target domain. 


\section{EXPERIMENT}
\subsection{Datasets}
\textbf{vKITTI to SemanticPOSS}. The synthetic dataset vKITTI \cite{gaidon2016virtual} contains 6 sequences of outdoor scenes in urban settings, where the point cloud are generated from the synthetic 2D depth images. The SemanticPOSS \cite{pan2020semanticposs} dataset was obtained in dynamic driving scenarios. It is composed of 6 sequences of scenes with a total of 2988 LiDAR scans. Therefore, there is a large gap in data distribution between the vKITTI and the real-world SemanticPOSS. For the domain adaptation scenario from vKITTI to SemanticPOSS, we select 6 semantic categories for domain adaptation: plants, building, road, traffic sign, pole, and car. The point clouds are sampled into blocks of 15m$\times$15m, and each block contains 4096 points.

\noindent\textbf{S3DIS to ScanNet}. The S3DIS \cite{armeni20163d} dataset is an indoor point cloud dataset containing 6 areas with 271 rooms and the ScanNet \cite{dai2017scannet} dataset contains 1513 indoor point cloud scenes annotated. For the domain adaptation scenario from S3DIS to ScanNet, we use 8 semantic categories for domain adaptation: floor, wall, window, door, table, chair, sofa, and bookshelf. Due to the sparsity and scene incompleteness of ScanNet, there is a huge domain gap between the datasets. We divide the point clouds into blocks of size 1.5m$\times$1.5m, and each block contains 8192 points.

\noindent\textbf{SemanticKITTI to nuScenes}. The SemanticKITTI \cite{behley2019semantickitti} dataset and the nuScenes \cite{fong2021panoptic} dataset are real-world datasets. However, the SemanticKITTI dataset is obtained by the 64-beam LiDAR scanner, while the nuScenes dataset is obtained by the 32-beam LiDAR scanner. Thus, there is a large gap of data sparsity in the SemanticKITTI-to-nuScenes domain adaptation scenario. We focus on the 10 categories for domain adaptation: car, bicycle, motorcycle, truck, other vehicle, pedestrian, drivable, sidewalk, terrain, and vegetation. The point clouds are sampled into blocks of 10m$\times$10m, and each block contains 4096 points.

\subsection{Implementation Details}
We use the official PyTorch implementation for PointWeb as our segmentation backbone. Stochastic Gradient Descent (SGD) optimizer is selected for training with the momentum 0.9 and the weight decay 0.0001, respectively. Also, we apply the weight decay to the learning rate, where the drop factor is 0.1 and the step size is 30. The initial learning rate for indoor and outdoor scenarios are 0.05 and 0.005. The capacity size $B$ of the memory bank is set to 16. The parameters $\lambda_1$ and $\lambda_2$ are set to 1.0 and 0.1. The margin $\alpha$ of the contrastive loss is set to 0.4. The values in k-NN for different levels are set to 1, 4, 16 and 64. To train and test our model, we use a single TITAN RTX GPU and the batch size is set to 4.

\begin{table}[h]
	\caption{The performance comparison of unsupervised domain adaptation methods. All the results are reported by mIoU.}
	\vspace{-14pt}
	\label{tab:1}
	\begin{center}
		\begin{tabular}{p{46 pt}<{\centering}|p{56 pt}<{\centering}|p{40 pt}<{\centering}|p{51 pt}<{\centering}}
			\hline
			\noalign{\smallskip}
			\multicolumn{1}{c}{\multirow{2}{*}{\small{Model}}}& \small{vKITTI to SemanticPOSS} & \small{S3DIS to ScanNet} & \small{SemanticKITTI to nuScenes}\\
			\noalign{\smallskip}
			\hline
			\noalign{\smallskip}
			\small{Supervised}	&\small{65.8}&\small{66.4}&\small{46.8}\\
			\noalign{\smallskip}
			\hline
			\noalign{\smallskip}
			\small{Source Only}	&\small{44.6}&\small{43.2}&\small{26.7}\\
			\small{MinEnt}		&\small{45.9}&\small{44.3}&\small{32.3}\\
			\small{MaxSquare}	&\small{46.3}&\small{43.6}&\small{32.8}\\
			\small{ADDA}		&\small{49.8}&\small{42.5}&\small{31.2}\\
			\small{PL}			&\small{51.0}&\small{46.4}&\small{29.8}\\
			\small3DGCA			&\small{47.1}&\small{43.1}&\small{33.7}\\
			\small{SQSGV2}		&\small{-}	&\small{-}&\small{10.1}\\
			\small{C\&L}		&\small{-}	&\small{-}&\small{31.6}\\
			\noalign{\smallskip}
			\hline
			\noalign{\smallskip}
			\small{Ours}				&\textbf{\small{54.9}}&\textbf{\small{53.8}}&\textbf{\small{37.3}}\\
			\noalign{\smallskip}
			\hline
			
		\end{tabular}
	\end{center}
	\vspace{-25pt}
\end{table}

\subsection{Performance Comparison}
We report the performance of point cloud semantic segmentation with mean Intersection-over-Union (mIoU). Tab.\ref{tab:1} shows the quantitative results of the comparison between our method with other domain adaptation methods. As shown in the table, our method achieves the highest performance, which shows the effective domain transferability of our method. Specifically, the Supervised means the model of PointWeb is trained on the target domain with semantic labels. The Source Only means the model trained with the source domain and directly tested at the target domain. Due to the domain gap, the performance of Source Only has a significant drop compared to the Supervised, which shows the necessity of unsupervised domain adaptation.

In order to verify the effectiveness of our method, we compare our method with a series of general unsupervised domain adaptation methods: MinEnt \cite{vu2019advent}, MaxSquare \cite{chen2019domain}, and ADDA \cite{tzeng2017adversarial}. For a fair comparison, these methods are reproduced with the same setting in our framework, where the hyperparameters are adjusted to obtain the best performance on all domain adaptation scenarios. The PL is the same Pseudo-Label training strategy in the \cite{jaritz2020xmuda} with our framework, which is a two-stage method with extra training cost. Furthermore, we introduce the geodesic correlation alignment used in \cite{wu2019squeezesegv2} into our segmentation framework to construct additional comparison method 3DGCA. It can be observed from the Tab. \ref{tab:1} that although the above methods can alleviate the domain discrepancy, they are not efficient for point cloud semantic segmentation. Especially with the global-level feature alignment methods, the model produces the confused semantic information in the feature space. In the S3DIS to ScanNet scenario, these methods even produce negative effects on domain adaptation.

Furthermore, we compare our method with the 3D unsupervised domain adaptation methods: SqueezeSegV2 (SQSGV2) \cite{wu2019squeezesegv2} and Complete \& Label (C\&L) \cite{yi2021complete}. Because these methods use different point cloud semantic segmentation backbones, we directly use the results of SQSGV2 and C\&L reported in \cite{yi2021complete} for comparison. Since SQSGV2 requires spherical projection and C\&L requires sequences of point cloud for the completion network, it is limited to reproduce in the vKITTI to SemanticPOSS and S3DIS to ScanNet domain adaptation scenarios. As shown in the Tab. \ref{tab:1}, compared with the above methods, our method achieves state-of-the-art performance on three domain adaptation scenarios.

\begin{figure}[t]
	\centering
	\includegraphics[width=1.0\linewidth]{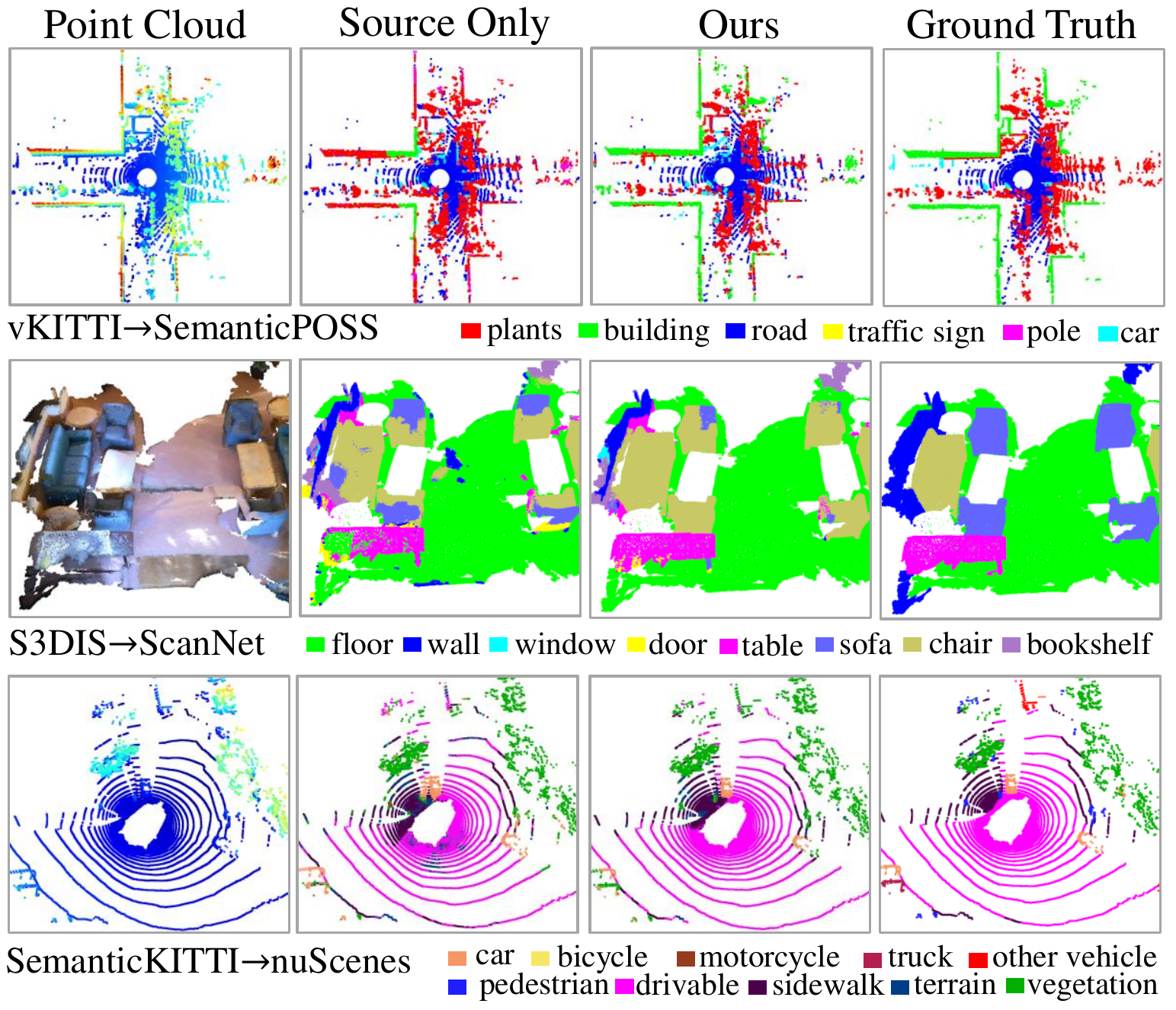}
	\vspace{-23pt}
	\caption{Visualization of point cloud semantic segmentation.}
	\label{fig:visual}
	\vspace{-18pt}
\end{figure}

\subsection{Ablation Studies and Analysis}

In order to further verify the effects of each module of our method and the effectiveness of the proposed assignment matrix based local feature loss, we conduct ablation studies on the vKITTI to SemanticPOSS scenario. 

As shown in Tab. \ref{tab:2}, we first report the performance improvement brought by each proposed loss, where the quantitative results of each loss can show its effective domain transferability. Particularly, the integration of the two losses can benefit the overall domain adaptation framework, and further improve the domain adaptation performance. It can be observed that our framework can benefit from a simple pseudo-label training strategy with additional 3.0\% improvement, and they play a complementary role in unsupervised domain adaptation. 

\begin{figure}[t]
	\centering
	\includegraphics[width=1.0\linewidth]{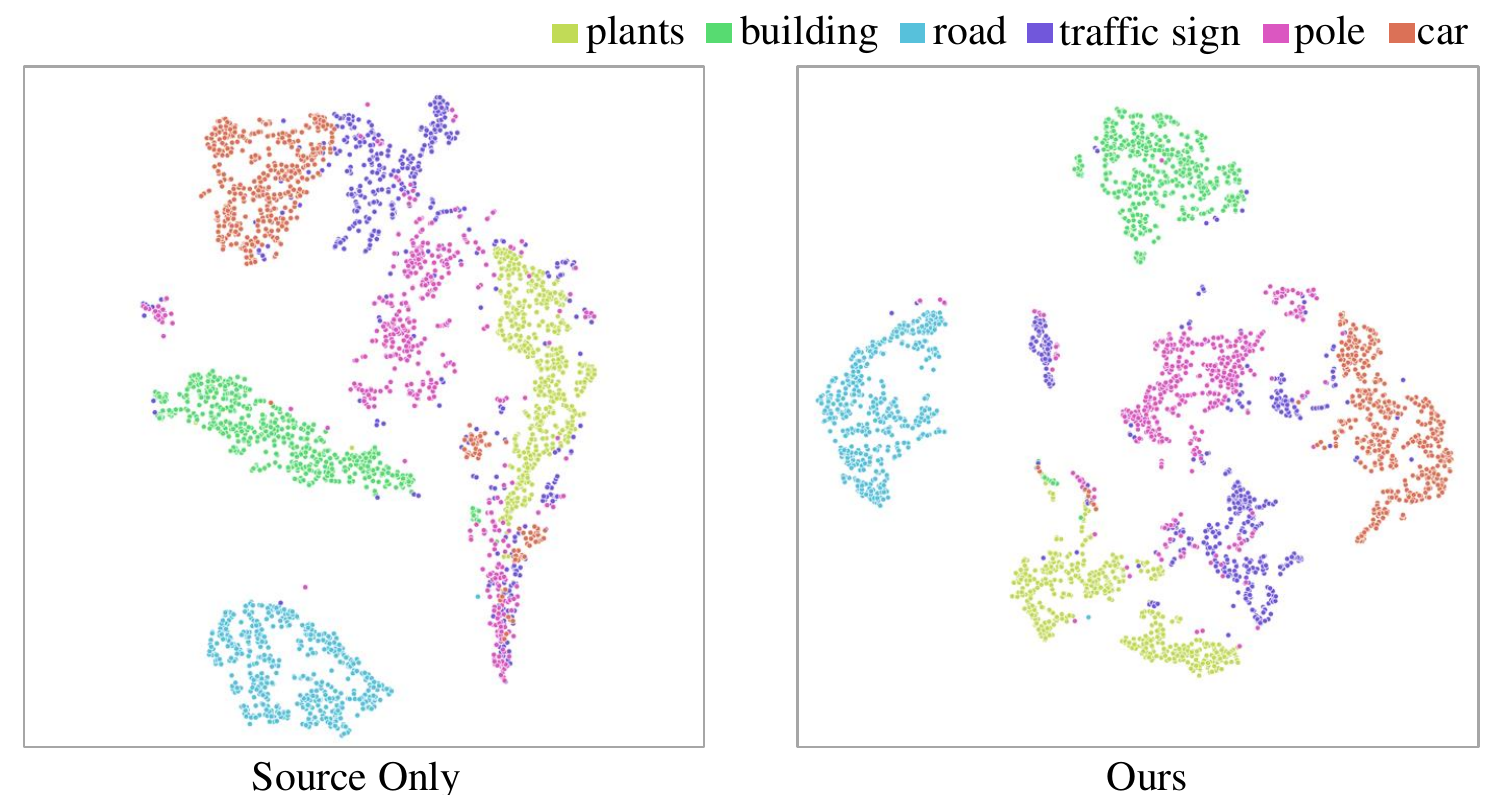}
	\vspace{-20pt}
	\caption{t-SNE visualization of features from the SemanticPOSS.}
	\label{fig:feature}
	\vspace{-20pt}
\end{figure}

Secondly, in order to verify that the feature distributions of different categories has been separated, we draw t-SNE \cite{van2008visualizing} visualization with the features from the target domain to show qualitative results. As shown in Fig. \ref{fig:feature}, our proposed method can effectively enhance the discrimination of features from the target domain.

Thirdly, we conduct the ablation study without using the assignment matrix named $L_{loc}$ w/o $A$. In this case, we use the mean features mentioned in Sec.\ref{sec:align} to represent the local feature graphs, and directly select the nearest neighbor from the memory bank to find the relationship between the graphs from the source domain and the target domain. Instead of using the assignment matrix for the alignment, we use the mean features to directly align the two graph features. As shown in Tab. \ref{tab:2}, the proposed assignment matrix based local feature loss can achieve a better performance.

At last, we show the visualization of point cloud semantic segmentation results to qualitatively illustrate the effectiveness of our method. It can be clearly observed in Fig. \ref{fig:visual}, compared with the Source Only, only a few noise predictions are produced in our method, which shows the proposed framework can effectively alleviate the domain gap problem and significantly improve the segmentation performance.

\begin{table}[h]
	\caption{Ablation study of adapting vKITTI to SemanticPOSS}
	\vspace{-17pt}
	\label{tab:2}
	\begin{center}
		\begin{tabular}{p{55 pt}<{\centering}|p{12 pt}<{\centering}p{12 pt}<{\centering}p{12 pt}<{\centering}p{12 pt}<{\centering}p{12 pt}<{\centering}p{12 pt}<{\centering}|p{18 pt}<{\centering}}
			\hline
			\noalign{\smallskip}
			\small{Model}& \small{\rotatebox{90}{plants}}& \small{\rotatebox{90}{building}}& \small{\rotatebox{90}{road}}&\small{\rotatebox{90}{trafficsign}}&\small{\rotatebox{90}{pole}}&\small{\rotatebox{90}{car}}&\small{mIoU}\\
			\noalign{\smallskip}
			\hline
			\noalign{\smallskip}
			\small{Source Only}				&\small{57.4}&\small{58.2}&\small{75.3}&\small{16.5}&\small{17.7}&\small{42.5}&\small{44.6}\\
			\small{$L_{loc}$ w/o $A$}		&\small{61.5}&\small{71.3}&\small{76.9}&\small{10.7}&\small{18.1}&\small{41.5}&\small{46.7}\\
			\small{$L_{loc}$}				&\small{62.1}&\small{73.6}&\small{79.9}&\small{9.7}&\small{25.1}&\small{44.2}&\small{49.1}\\
			\small{$L_{con}$}				&\small{60.0}&\small{72.0}&\small{78.1}&\small{15.4}&\small{27.1}&\small{45.8}&\small{49.7}\\
			\small{$L_{loc}$+$L_{con}$}		&\small{63.2}&\small{74.8}&\small{81.9}&\small{12.6}&\small{28.8}&\small{50.0}&\small{51.9}\\
			\small{$L_{loc}$+$L_{con}$+PL}	&\small{\textbf{63.9}}&\small{\textbf{76.9}}&\small{\textbf{84.1}}&\textbf{\small{16.6}}&\small{\textbf{36.4}}&\small{\textbf{51.5}}&\small{\textbf{54.9}}\\
			\noalign{\smallskip}
			\hline
		\end{tabular}
	\end{center}
	\vspace{-15pt}
\end{table}

\section{CONCLUSION}

In this paper, we proposed an unsupervised domain adaptive point cloud semantic segmentation framework based on feature graph matching. With the proposed assignment matrix based local feature loss and category-guided contrastive loss, we can align the local-level feature distributions of the source domain and the target domain more accurately in a meticulous way and guide the segmentation model to learn discriminative features on the target domain. Extensive experiments on different synthetic-to-real and real-to-real domain adaptation scenarios have demonstrated the superiority of our method.

\bibliographystyle{IEEEtran}
\bibliography{IEEEabrv,reference}

\end{document}